\documentclass[runningheads]{llncs}
\usepackage[T1]{fontenc}
\usepackage{url}            
\usepackage{booktabs}       
\usepackage{amsmath}

\usepackage{hyperref}
\usepackage{cleveref}
\usepackage{amsfonts}       
\usepackage{nicefrac}       
\usepackage{microtype}      
\usepackage{comment}
\usepackage{tabularx}
\usepackage{graphicx}
\usepackage{array,multirow}
\usepackage{pifont}
\graphicspath{{Styles/figures/}{figures/}}
\usepackage{caption}
\usepackage{wrapfig}
\usepackage[table]{xcolor} 
\usepackage[misc]{ifsym}
\newcommand{\corr}{(\Letter)}
\usepackage{mwe}
\definecolor{slatecolor}{rgb}{1, 0.95, 0.9}
\definecolor{darkgreen}{rgb}{0.0, 0.5, 0.0}

\hypersetup{
    colorlinks = true,
    linkcolor = {red},
    citecolor = {darkgreen},
    urlcolor = {darkgreen},
    allbordercolors = {white}
} 
\begin{document}

\title{Temporal receptive field in dynamic graph learning: A comprehensive analysis}

\titlerunning{Temporal receptive field analysis}

\author{Author information scrubbed for double-blind reviewing}

\author{Yannis Karmim \inst{1} \and 
Leshanshui Yang \inst{3,4} \and
Raphaël Fournier S'niehotta \inst{1} \corr \and
Nicolas Thome \inst{2} \and
Cl\'ement Chatelain\inst{5} \and
S\'ebastien Adam\inst{3}  
}


\authorrunning{Karmim et al.}

\institute{Conservatoire National des Arts et Métiers, Paris 75003, France \email{\{yannis.karmim,fournier\}@cnam.fr}
\and
Sorbonne Université Sciences, Paris 75005, France \\ \email{nicolas.thome@isir.upmc.fr} \and
Univ Rouen Normandie, LITIS UR 4108, F-76000 Rouen, France \email{clement.chatelain@insa-rouen.fr}\and
Saagie, 72 Rue de la République, 76140 Le Petit-Quevilly, France  \email{leshanshui.yang@saagie.com}\and 
INSA Rouen Normandie, LITIS UR 4108, F-76000 Rouen, France \email{sebastien.adam@univ-rouen.fr} 
}

\maketitle              

\begin{abstract}
Dynamic link prediction is a critical task in the analysis of evolving networks, with applications ranging from recommender systems to economic exchanges. However, the concept of the temporal receptive field, which refers to the temporal context that models use for making predictions, has been largely overlooked and insufficiently analyzed in existing research. In this study, we present a comprehensive analysis of the temporal receptive field in dynamic graph learning. By examining multiple datasets and models, we formalize the role of temporal receptive field and highlight their crucial influence on predictive accuracy. Our results demonstrate that appropriately chosen temporal receptive field can significantly enhance model performance, while for some models, overly large windows may introduce noise and reduce accuracy. We conduct extensive benchmarking to validate our findings, ensuring that all experiments are fully reproducible. Code is available at \href{https://github.com/ykrmm/BenchmarkTW}{https://github.com/ykrmm/BenchmarkTW}.
\end{abstract}

\section{Introduction}

Dynamic graphs play an essential role in modeling evolving interactions between entities across various domains, from social networks to computational biology \cite{Fan2019,jumper2021highlyalphafold}. The task of link prediction on such graphs is essential, with numerous applications including forecasting user behavior in recommendation systems, predicting financial transactions, and identifying potential collaborations in academia.  Dynamic graphs in these domains are often represented as either a sequence of static graphs captured at regular intervals, referred to as Discrete Time Dynamic Graphs (DTDG) \cite{skarding2021foundationssurvey,yang2024dynamicsurvey}, or as Continuous Time Dynamic Graphs (CTDG) which capture interactions as they occur over continuous time \cite{poursafaei2022towardsprediction,rossi2020tgn:graphs,wang2021inductivewalks,xu2020tgat:graphs}.

Models that operate on Discrete Time Dynamic Graphs (DTDGs) usually employ a combination of Message-Passing Graph Neural Networks (MP-GNNs) to capture spatial dependencies and Recurrent Neural Networks (RNNs) to capture temporal dependencies \cite{You2022ROLANDGL,pareja2019evolvegcn:graphs,sankar2018dysat:networks,model_chen2022gclstm}. 
By capturing the evolution of graph structures over time and integrating this dynamic information, these models predict future links with greater accuracy and robustness, addressing the inherent complexities of evolving networks.

However, real-world dynamic graph data come in diverse forms and span different numbers of snapshots. The critical factor is the "length" of dependencies to consider in terms of the number of time steps. This variability poses significant challenges for modeling and prediction, as the number of relevant snapshots can greatly affect the performance of predictive models \cite{liu2021anomalytransformer,intro_tw_param_VDGCNeT_ZHENG2023110676,Intro_tw_worse_biparva2024todyformer}.

Some models propose to use a temporal window as a hyper-parameter to address this issue \cite{liu2021anomalytransformer}. As illustrated in \cref{fig:dynamic_link_prediction}, a temporal window defines the span of past data considered for making predictions. Results have shown that in certain cases, a large temporal window can introduce noise and degrade performance. For example, some models like TADDY \cite{liu2021anomalytransformer} or Todyformer \cite{Intro_tw_worse_biparva2024todyformer} demonstrate that excessive temporal data can lead to overfitting or irrelevant information overshadowing crucial patterns.

Beyond these marginal hyper-parameter searches in a limited number of models \cite{liu2021anomalytransformer}, there is a notable lack of a comprehensive analysis on the impact of the temporal window across the wide range of datasets used by the "learning on dynamic graphs" community. This gap calls for a systematic evaluation to understand how different temporal contexts affect dynamic link prediction performance.

In this work, we conduct a comprehensive analysis of the impact of the temporal receptive field on dynamic link prediction performance. We evaluate multiple Discrete Time Dynamic Graph Neural Networks (DTDGNNs) across diverse datasets, examining how different temporal receptive fields influence predictive accuracy. Our contributions are as follows:

\begin{itemize}
    \item \textbf{Temporal receptive field analysis}: We formalize the concept of temporal receptive fields in dynamic graph learning and examine their influence on predictive accuracy and model performance across various scenarios.
    \item \textbf{Model and dataset evaluation:}: By comparing the performance of multiple DTDGNNs on a diverse set of datasets, we identify specific contexts in which different temporal window lengths optimize model performance. This analysis provides insights into which datasets benefit most from dynamic information and guides the selection of optimal temporal receptive field.
    \item \textbf{Benchmarking and reproducibility}: We conduct extensive benchmarking by running various DTDG models across multiple datasets. All the experiments in this work are fully reproducible. We provide full access to the code, models, configuration files, and datasets. Our resources ensure that other researchers can easily integrate new datasets and models, facilitating rapid evaluation of the need for temporal context in dynamic link prediction.
\end{itemize}


\section{Context and related work} \label{sec_preliminaries}

This section reviews a few key concepts relevant to the dynamic link prediction task with discrete time dynamic graph neural networks.

\subsection{Discrete Time Dynamic Graph Link Prediction}

\begin{figure}[!ht]
  \centering  \includegraphics[width=.8\textwidth]{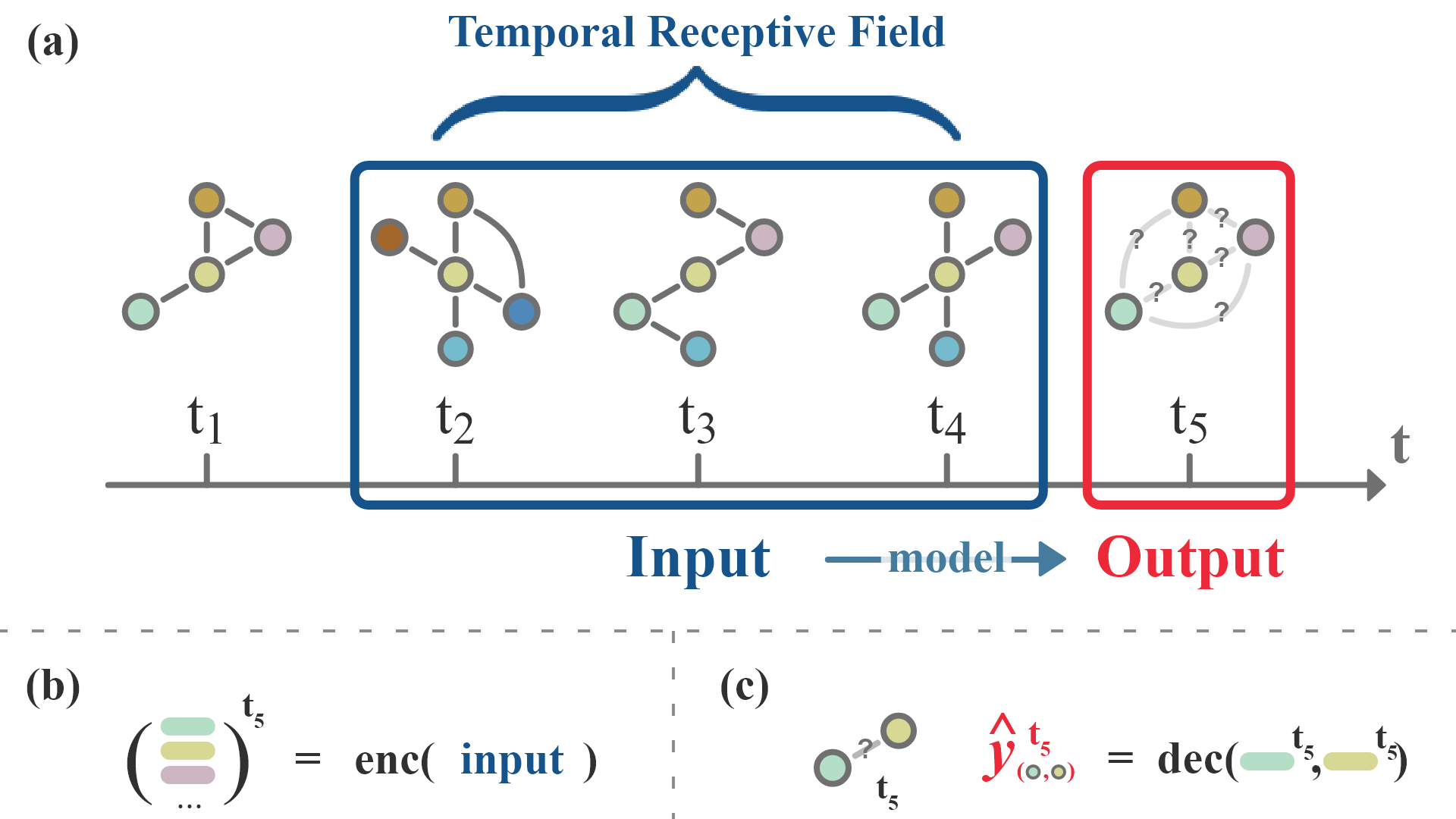}
   \caption{
    \textbf{(a)} Dynamic Link Prediction task on a Discrete Time Dynamic Graph: In this figure, the model takes snapshots from $t_2$ to $t_4$ as input and predicts the existence of edges at time $t_5$. The number of time steps in which the model can capture information is called the size of the receptive field, denoted as $\tau$.\\   
    \textbf{(b)} When predicting, for example at $t_5$, the encoder of the model computes node representation vectors based on the input.\\
    \textbf{(c)} For each edge to be predicted, the decoder of the model computes the relevant node representations to obtain the probability of the existence of the edge. The figure shows the edge between the green and yellow nodes at time $t_5$ as an example.
    }
   \label{fig:dynamic_link_prediction}
\end{figure}

A Discrete Time Dynamic Graph (DTDG) can be viewed as a sequence of static graphs (called snapshots) as shown in Equation \ref{eq:dtdg} and Figure~\ref{fig:dynamic_link_prediction} (a). Each snapshot $G^t$ is represented by a set of nodes $V^t$ and a set of edges $E^t$, with optional $\mathbf{X}_v^t$ or $\mathbf{X}_e^t$ representing node or edge attributes \cite{skarding2021foundationssurvey,survey_zaki2016comprehensive}. $N^t = |\mathcal{V}^t|$ and $M^t = |\mathcal{E}^t|$ indicate respectively the numbers of nodes and edges in snapshot $G^t$.

\begin{equation} \label{eq:dtdg}
    \textit{DTDG} = (G^1, G^2, ..., G^T)
\end{equation}

The dynamic link prediction task aims to predict the existence of an edge $e_{u,v}^{t+1}$ between two nodes $u$ and $v$ at time $t+1$ using past information from a time-window of size $\tau$ $ \{G^{t-\tau +}, \ldots, G^{t}\} $.
A model learns to represent nodes and/or edges with the first $T_{train}$ snapshots of the DTDG.
Then, for each snapshot $G^t$ in the subsequent $T_{test}$ time steps, the model predicts the probability of the existence of the edges in the future time step $t+1$.

\subsection{Discrete Time Dynamic Graph Neural Networks}

To address DTDG tasks such as Dynamic Link Prediction, encoder-decoder structures are widely used in Discrete Time Dynamic Graph Neural Networks (DTDGNNs) \cite{model_li2017dcrnn,model_chen2022gclstm,model_king2023euler}. 
The encoder transforms input signals such as attributes and structure of the graph into latent representations. This process is also known as representation learning. The node representations are denoted as \( \mathbf{Z}_v^t \in \mathbb{R}^{N^t \times d} \), where \( d \) represents the size of the vector, as shown in figure \ref{fig:dynamic_link_prediction} (b). Edges can be encoded in the same way \cite{EdgeEmb_renton,EdgeEmb_pmlr_v129_yang20a}.  
The decoder, coming after the encoder in a neural network, computes latent representations to obtain predictions for downstream tasks \cite{survey_hamilton2020graph}, as shown in figure \ref{fig:dynamic_link_prediction} (c).

To handle both temporal and structural information, the DTDG encoder typically combines static graph encoders $ f_G(\cdot) $ with temporal encoders $ f_T(\cdot) $ \cite{yang2024dynamicsurvey}. 
Graph encoders commonly rely on Graph Neural Networks (GNN) \cite{survey_wu2020comprehensive}, Random Walk-based methods \cite{model_graph_grover2016node2vec}, or Graph Transformers \cite{survey_min2022transformer} for propagating node representations and capturing structural information within a snapshot. 
Temporal encoders typically leverage Recurrent Neural Networks (RNN) \cite{hochreiter1997longmemory}, Temporal Convolutional Networks (TCN) \cite{model_temporal_oord2016wavenet}, or Transformers \cite{vaswani2017attentionneed} for capturing temporal patterns among multiple time steps.

\begin{figure}[!]
  \centering
  \includegraphics[width=\textwidth]{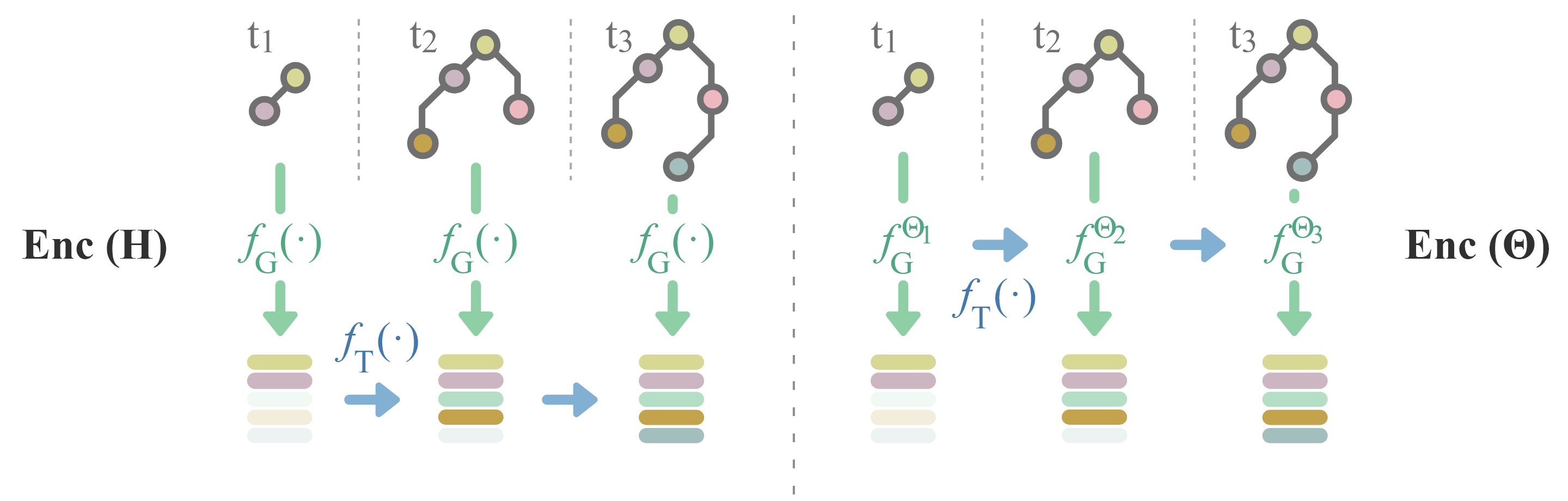}
  \caption{Two categories of Discrete Time Dynamic Graph Neural Networks (DTDGNNs). Left: Sequentially encoding the hidden states \textbf{H} of each snapshot across time with a temporal encoder $f_T(\cdot)$. Right: Sequentially encoding the parameters $\mathbf{\Theta}$ of the graph encoder $f_G(\cdot)$ across time with a temporal encoder $f_T(\cdot)$.}
   \label{fig:dtdgnns}
\end{figure}

From the perspective of how to encode information across multiple snapshots, DTDGNNs fall into two main paradigms: Sequentially Encoding Hidden states and Sequentially Encoding Model Parameters, denoted as $\mathbf{Enc(H)}$ and $\mathbf{Enc(\Theta)}$ respectively, as shown in figure \ref{fig:dtdgnns}.

The $\mathbf{Enc(H)}$ paradigm uses the graph encoder $ f_G (\cdot)$ to encode each snapshot as a static graph, and then uses the temporal encoder $ f_T (\cdot) $ to encode temporal information across snapshots, as shown in equation \ref{eq:ench}. The two modules can be combined in either a stacked or integrated structure. 
In stacked structures, $f_G$ and $f_T$ layers are stacked one or several times. 
For example, \textbf{DySAT} \cite{sankar2018dysat:networks} successively uses structural self-attention \cite{vaswani2017attentionneed} and temporal self-attention to encode graph and temporal information; 
\textbf{STGCN} \cite{model_yu2017stgcn} uses stacked (TCN, GNN, TCN) layers as the basic module and repeats it several times.

$\mathbf{Enc(H)}$ with integrated structure takes another approach by incorporating $f_G(\cdot)$ into $f_T(\cdot)$. 
Given that $f_T(\cdot)$ has linear projection \cite{hochreiter1997longmemory} or convolution modules \cite{model_temporal_oord2016wavenet} to handle the node features, this approach aggregates neighboring features by replacing these modules with $f_G(\cdot)$. 
Typical models include \textbf{GC-LSTM} \cite{model_chen2022gclstm} replacing the linear layer in LSTM \cite{hochreiter1997longmemory} with GCN \cite{kipf2016semi-supervisednetworks}. 

\begin{eqnarray}
    \label{eq:ench}
    \mathbf{H'}^k = f_G(\mathbf{A}^k, \mathbf{H}^k), \quad k \in [t-\tau+1, t] \\ \nonumber
    \mathbf{H}^{t+1} = f_T(\mathbf{H'}^{t-\tau+1:t})
\end{eqnarray}

Another paradigm $\mathbf{Enc(\Theta)}$ lets the parameters of $f_G(\cdot)$ keep evolving. This paradigm uses $f_T(\cdot)$ to encode the parameters of $f_G(\cdot)$ across snapshots, allowing each snapshot to be encoded in a more adaptive way. 
\textbf{EvolveGCN} (EGCN) \cite{pareja2019evolvegcn:graphs} suggests dynamically updating the GCN \cite{kipf2016semi-supervisednetworks} weight parameters. As shown in equation \ref{eq:enctheta}, the graph encoder parameters $ \mathbf{\Theta}_{f_G}^{t+1} $ in each snapshot are predicted by a temporal encoder.

\begin{eqnarray} \label{eq:enctheta}
    \mathbf{\Theta}_{f_G}^{t+1} = f_T(\mathbf{\Theta}_{f_G}^{t}, \mathbf{H}^{t}) \quad \mathbf{H}^{t+1} = f_G(\mathbf{A}^{t}, \mathbf{H}^{t}; \mathbf{\Theta}_{f_G}^{t+1})
\end{eqnarray}

The diverse model architectures demonstrate how existing research incorporates temporal contexts when extending static graph encoders to dynamic graphs. This opens up possibilities for various tasks and applications of DTDG.

\section{Temporal Receptive Field in Dynamic Graph Representation Learning} \label{sec:temporalrf}

Capturing temporal patterns plays an important role in Dynamic Graph Representation Learning. Toward better temporal encoding, existing research has primarily focused on structural optimizations of $ f_T (\cdot)$, such as incorporating attention mechanisms \cite{sankar2018dysat:networks} or gating mechanisms \cite{model_li2017dcrnn,model_yu2017stgcn}. However, there has been limited discussion of the temporal receptive field of DTDGs.

We introduce the concept of the \textbf{temporal receptive field}. The temporal receptive field refers to the temporal information or contexts in the data that a model can capture and use when processing sequential data \cite{ref_receptive_field_wang2020,temporalcontext_peddinti2015time}. Different neural network architectures handle temporal contexts in various ways.

Recurrent Neural Networks (RNN) \cite{hochreiter1997longmemory,model_temporal_chung2014gru} process sequential data through a recursive structure, where the output of each time step depends not only on the current input but also on the hidden state from the previous time step. Due to this recursive nature, each step connects to the entire time steps in the given input, theoretically making its temporal receptive field encompass all input time steps. 
Practically, vanilla RNNs are trained using the Backpropagation Through Time algorithm that unfolds the network, making it deeper and subject to the vanishing gradient issue \cite{temporalcontext_pascanu2013difficulty}. Therefore, the unfolded network is generally limited to a reduced number of time steps, thus limiting their temporal receptive field. LSTM and GRUs partially compensate for this limitation by adding an explicit memory cell in their units, but can still suffer from the vanishing gradient.

Among non-RNN structures, common temporal encoders include Temporal Convolutional Networks (TCNs) and Transformers. TCNs apply causal convolution \cite{model_temporal_oord2016wavenet} along time steps to process sequential data. Transformers utilize self-attention mechanisms \cite{vaswani2017attentionneed} to process sequential data, allowing each input position to interact with all other positions in the sequence. The size of their receptive field $\tau$ is explicitly determined by model hyper-parameters.

Intuitively, a larger temporal receptive field contains richer past information for the model to capture. However, in practice, RNNs suffer from the problem of vanishing or exploding gradients \cite{temporalcontext_pascanu2013difficulty}. More complex RNN models such as LSTM and GRU also have problems with the effectiveness of memory over long sequences \cite{temporalcontext_bai2018empirical,LongMemory_pmlr_v119_zhao20c}. Empirically, some models indicate that incorporating more time steps in dynamic graphs does not necessarily yield better performance \cite{liu2021anomalytransformer,Intro_tw_worse_biparva2024todyformer}.

To our knowledge, the influence of the temporal receptive field size on model performance on DTDGs remains an unexplored area. Many analyses of the temporal receptive field of the inputs/outputs in DTDGs focus on hyper-parameter search \cite{liu2021anomalytransformer,model_king2023euler,fan2022gnn,intro_tw_param_VDGCNeT_ZHENG2023110676}, highlighting the necessity of this research. This emphasizes the importance of a benchmark in this field to investigate how much temporal context is necessary to perform effectively on dynamic link prediction tasks, and extend to generalized DTDG representation learning.

\section{Experiments}

The dynamic link prediction task is often modeled as a binary classification problem, where truly existing edges labeled as 1 and non-existent edges labeled as 0 \cite{pareja2019evolvegcn:graphs,model_chen2022gclstm,model_king2023euler,sankar2018dysat:networks}.
As a standard practice, existing studies use the "negative sampling" method when evaluating model performance \cite{pareja2019evolvegcn:graphs}. This method samples $M^{t+1}$ non-existent edges and mixes in $M^{t+1}$ truly existing edges at $t+1$ to formulate a balanced binary classification task. The average precision is therefore widely used to evaluate the performance of the model. This is also the case in our experiments.

The goal of these experiments is to address the following questions: (1) How does the temporal receptive field influence results in dynamic link prediction (\cref{sec:exp_perf}) ?  (2) Which datasets require the broadest temporal receptive field (\cref{sec:expdataanalysis}) ?  (3) What is each model's capacity to capture dynamic information (\cref{sec:model_analysis}) ?  Additionally, these experiments provide a rigorous benchmarking of several models across a wide range of datasets used by the dynamic graph learning community. At the end of this experimental section, we also discuss the implications of the results for DTDGNN models and suggest future research directions (\cref{sec:discussion}). 
\subsection{Baselines and Datasets}
\begin{table}[!ht]
\setlength\tabcolsep{4.5pt}

  \centering
  \footnotesize 
  \caption{Statistics of datasets used in our experiments.}
  \begin{tabular}{l|lllll} 
    \toprule
    Datasets & Domains & Nodes & Links & Snapshots & Duration \\
    \midrule
    CanParl & Politics & 734 & 74,478 & 14 & 14 years\\
    USLegis & Politics & 225 & 60,396 & 12 & 12 congresses\\
    Trade & Economics & 255 & 507,497 & 32 & 32 years \\
    UNVote & Politics & 201 & 1,035,742 & 72 & 72 years \\
    UCI-Message & Social & 1,899 & 59,835 & 88 & 196 days \\
    AS733 & Router & 6,628  & 13,512 & 30 & 86 days \\
    Enron & Mail & 184 & 790  & 11 & 3 years\\
    Colab & Citations & 315 & 943 & 10 & 9 years \\
    Bitcoin-OTC & Trust Networks & 5,881 & 35,592  & 136 & 5 years\\
    Bitcoin-Alpha & Trust Networks & 3,783 & 24,186 & 136 & 5 years\\
    \bottomrule
  \end{tabular}
  \label{tab:datasets}
\end{table}
To answer these questions, we selected ten discrete dynamic graph datasets (\cref{tab:datasets}) commonly used by the dynamic graph learning community \cite{pareja2019evolvegcn:graphs,sankar2018dysat:networks,poursafaei2022towardsprediction}. These graphs are diverse, ranging from social networks to economic and political networks. The datasets CanParl, USLegis, Trade, and UNVote are sourced from the dynamic link prediction evaluation paper \cite{poursafaei2022towardsprediction}. The UCI-Message, AS733, Bitcoin OTC, and Bitcoin Alpha datasets were used by the EGCN model \cite{pareja2019evolvegcn:graphs} and are also part of the Stanford database \cite{snapnets}. We also included the Enron email dataset and the scientific collaboration dataset Colab from \cite{snapnets}.
An initial observation is the disparity in the lifespan of these DTDGs, ranging from a few days for the AS733 router network to almost a century for the UNVote dataset. We provide more details of each dataset in section B of the appendix.

\subsection{Enhancing dynamic link prediction with an optimal temporal receptive field} \label{sec:exp_perf}
\begin{figure}[!h]
    \centering
    \includegraphics[width=\linewidth]{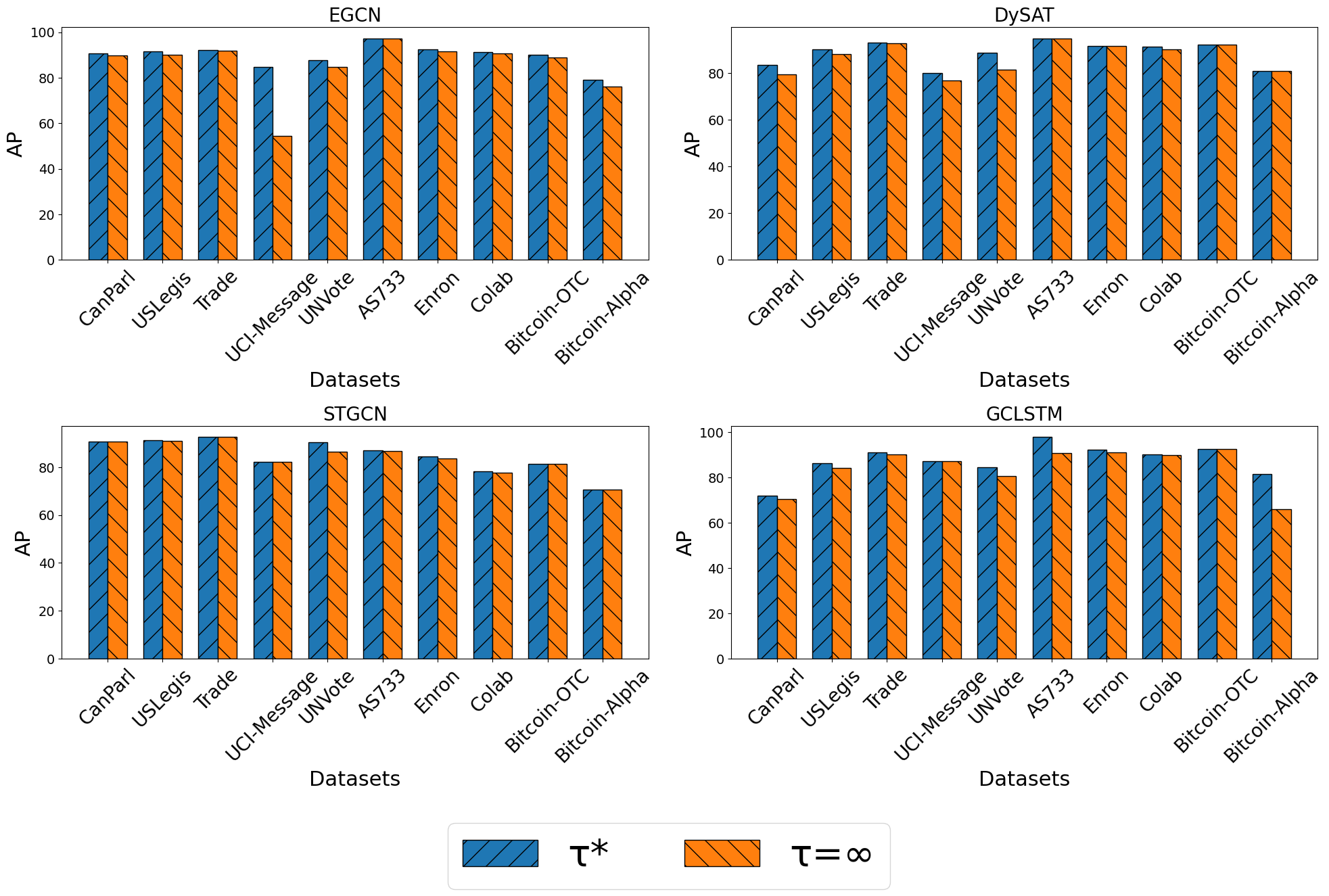}
    \caption{Model performance in average precision, with an optimal temporal receptive field $\tau^*$ vs. all temporal information $\tau_\infty$. }
    \label{fig:tau_opt}
\end{figure}
This subsection and results address the question: \textit{(1)How does the temporal receptive field influence results in dynamic link prediction?}
Our first results in \cref{fig:tau_opt} demonstrate that adapting the temporal receptive field $\tau$ can significantly enhance model performance. The first part of the table is the score of models using all the temporal information, with a temporal receptive field of $\tau_\infty$. This is the standard implementation of existing DTDGNNs models \cite{sankar2018dysat:networks,model_yu2017stgcn,pareja2019evolvegcn:graphs,model_chen2022gclstm}. To find the optimal value of $\tau$, we conducted a parameter search within a range from 1 to the number of snapshots. The results for the optimal $\tau$ are shown in the second part of the table $\tau^*$. The optimal $\tau$ value is indicated in parentheses following the AP score.
Using the optimal temporal receptive field $\tau^*$ generally improves performance compared to using all temporal information $\tau_{\infty}$. For instance, \textbf{EGCN} shows an average gain of +3.40 points in AP, while \textbf{GCLSTM} improves by +3.24 points, \textbf{DySat} +1.80, \textbf{Edgebank} +1.92 and 0.58 for \textbf{STGCN}. These findings highlight the importance of selecting an appropriate $\tau$ tailored to the models and the dataset's characteristics.

For most datasets, using the optimal temporal receptive field $\tau^*$ leads to marked improvements in performance compared to using all temporal information $\tau_{\infty}$. This indicates that carefully selecting a temporal window size can enhance predictive accuracy significantly. We present more detailed results in \cref{tab:all_temp_vs_opt_tau} of section A of the appendix.

\subsection{Analysis of temporal receptive field size and dataset characteristics} \label{sec:expdataanalysis}
\begin{table}[!h]
    \centering
    \begin{tabular}{c | c | lllll}
    \toprule
        $\tau$ & Datasets & EGCN \cite{pareja2019evolvegcn:graphs} & DySat \cite{sankar2018dysat:networks} & GCLSTM \cite{model_chen2022gclstm} & STGCN \cite{model_yu2017stgcn} & EdgeBank \cite{poursafaei2022towardsprediction}\\
        \midrule
         \multirow{12}{*}{$\tau_{\infty} $} & CanParl & 89.70 & \textbf{79.44} & \textbf{70.47 }& 90.64 & \textbf{54.48}\\
         & USLegis & 90.02 & \textbf{88.36} & \textbf{84.42} & \textbf{90.95} & 57.75\\
         &Trade & 91.95& 92.78 & 90.15 & 92.62 & 65.94\\
         & UCI-Message &54.33 & \textbf{77.04 }& \textbf{87.22}  & \textbf{82.14} & \textbf{76.28} \\
         &UNVote & 84.76 & 81.60 & \textbf{80.75} & 86.49 & 61.17\\
         &AS733 & 97.17 & \textbf{94.88} & 90.86 & 86.66 & \textbf{97.21}\\
         &Enron & 91.58 & \textbf{91.84} & \textbf{91.26} & 83.73 & \textbf{79.35}\\
         &Colab & 90.69 & 90.20 & \textbf{89.99} &\textbf{77.68} & \textbf{77.02}\\
         &Bitcoin-OTC & 89.04 &  \textbf{92.23} &\textbf{92.65} & \textbf{81.29} & \textbf{52.18}\\
         &Bitcoin-Alpha & \textbf{76.19} & \textbf{81.10} & 65.91 & \textbf{70.72} & \textbf{57.43}\\
         \midrule
         \multirow{12}{*}{$\tau_1 $} & CanParl & \textbf{90.56} & 73.02 & 53.03 & 90.65 & 51.37\\
         & USLegis & \textbf{90.74} & 83.94 & 73.33 & 90.25 & 73.46\\
         &Trade & \textbf{92.12} & \textbf{93.07} & \textbf{90.76} & \textbf{92.70} & \textbf{82.42}\\
         & UCI-Message & \textbf{84.85} & 66.91 & 85.22 & 81.04 & 55.06\\
         &UNVote & \textbf{86.96} & \textbf{83.93} & 74.43 &\textbf{87.89} & \textbf{97.30} \\
         &AS733 & \textbf{97.37} & 94.57 & \textbf{98.03} & \textbf{86.94} & 82.15 \\
         &Enron & \textbf{92.45} & 91.28 & 89.29 & \textbf{84.49 }& 71.53 \\
         &Colab & \textbf{90.86} & \textbf{91.39} & 89.44 & 76.89 & 70.34 \\
         &Bitcoin-OTC & \textbf{89.99} & 60.32 & 87.55 & 79.99 & 50.08 \\
         &Bitcoin-Alpha & 73.51 & 54.78 & \textbf{81.70} & 67.00 & 51.33 \\
         \bottomrule
    \end{tabular}
    \caption{Average precision of DTDG models on multiple datasets using all temporal information $\tau_\infty$ vs using no dynamic information $\tau_1$.}
    \label{tab:all_temp_vs_tau1}
\end{table}
This subsection addresses the question (2): \textit{Which datasets require the broadest temporal receptive field?}
In the \cref{tab:all_temp_vs_tau1}, we show that for some datasets, using the last snapshot alone (i.e $\tau=1$) yields high predictive performance across all models. For the \textbf{UNVote} dataset, most models also achieve better predictions without using dynamic information. This is particularly notable with the non-parametric \textbf{EdgeBank} baseline, where predicting the same links as in the previous time step almost reaches perfect results. This can be explained by the nature of the dataset. The \textbf{UNVote} dataset documents roll-call votes conducted in the United Nations General Assembly. Whenever two nations cast a "yes" vote for an item, a link is established between them. Often, the policies of two countries are aligned within a short time frame, so looking at the relationships from 70 years ago is not particularly helpful. For the \textbf{Trade} dataset, we observe a similar phenomenon for all models. The \textbf{Trade} dataset covers trade in food and agriculture products between 181 nations over more than 30 years, with links reflecting the cumulative sum of normalized import or export values. Short-term trading patterns are more relevant, making older data less useful for predictions.

In contrast, for the \textbf{Bitcoin-OTC} and \textbf{Bitcoin-Alpha} datasets, users' reputations must be tracked over time to avoid fraudulent transactions, making long-term interactions crucial. In the \textbf{UCI-Message} dataset, although the duration is only 196 days, social interactions and message exchanges evolve rapidly, necessitating a broader temporal context to capture meaningful patterns and trends.\\
However, the \cref{tab:all_temp_vs_tau1} also shows that performance does not solely depend on the datasets but also on the model's ability to capture long-term dynamic information. For example, with {EGCN}, we observe that performance with a receptive field of $\tau_1$ is almost always better. This leads us to the next subsection and the need to analyze the different models in more depth.
\subsection{Analysis of the models' ability to capture dynamic information.} \label{sec:model_analysis}
This subsection addresses the question 3: \textit{What is each model's capacity to capture dynamic information?}
To do so, we propose two analyses. 
\begin{figure}[!h]
    \centering
    \includegraphics[width=\linewidth]{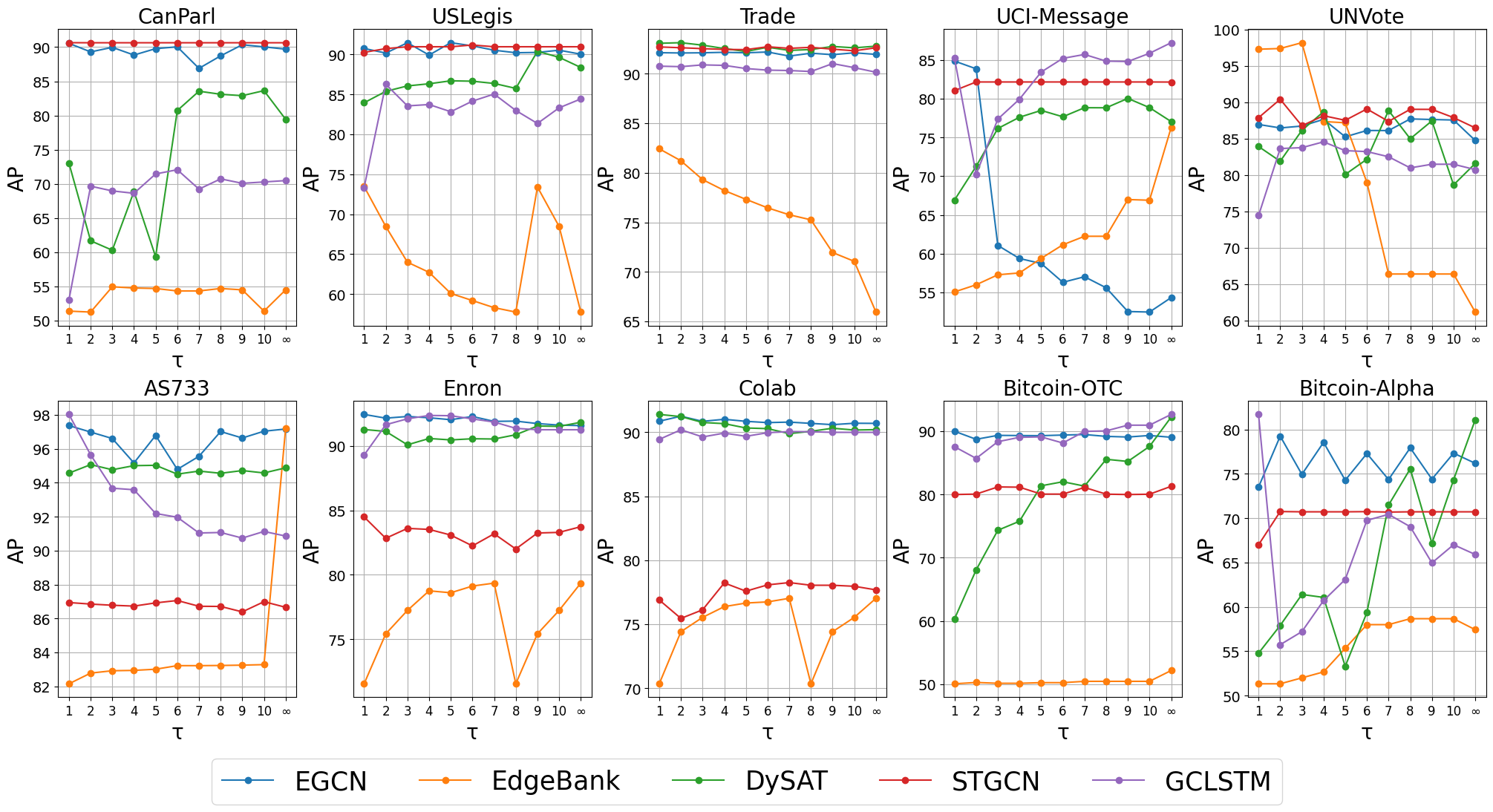}
    \caption{Average precision (AP) scores of various DTDG models across multiple datasets, shown as a function of the temporal receptive field $\tau$.  A value of $\tau_\infty$ represents the use of all temporal information.}
    \label{fig:tw}
\end{figure}
\subsubsection{a.) Performance of models in link prediction with varying temporal receptive fields $\tau$ (\cref{fig:tw})}
The figure \cref{fig:tw} presents the average precision scores of various DTDG models across multiple datasets as a function of the temporal receptive field $\tau$.
For most datasets, the performance of the \textbf{EGCN} model remains relatively better with smaller value of $\tau$, with a noticeable drop for larger temporal receptive field in some cases, such as Trade and UNVote. This suggests that \textbf{EGCN} is not able to capture long-term dynamic information.
The \textbf{DySAT} model shows more variability across datasets, but performs better with larger temporal receptive field. For USLegis and UNVote, performance improves a lot with larger temporal windows, indicating the ability of capturing dynamic information.

\textbf{GCLSTM} generally benefits from larger temporal receptive field, especially in the UCI-Message dataset, in \cref{sec:expdataanalysis} we underline the importance of long temporal context in this dataset.
\textbf{STGCN} is relatively stable as the temporal receptive field increases. Except UNVote and Trade for the reason that we explain in \cref{sec:expdataanalysis}, \textbf{EdgeBank} is better when it consider all temporal information. 

\subsubsection{b.) Correlation between the number of snapshots and the AP score in dynamic link prediction (\cref{tab:correlation_duration_performance})}
To evaluate the models' ability to capture long-term information, we calculated the correlation between the difference in performance at $\tau_\infty$ and $\tau_1$ and the number of snapshots. This correlation analysis in \cref{tab:correlation_duration_performance} reveals that models such as \textbf{DySAT} and \textbf{STGCN }exhibit a positive correlation with the total number of snapshots, demonstrating their ability to capture long-term temporal information. Conversely, \textbf{EGCN} and \textbf{GCLSTM }show negative correlations, indicating a reduced capacity to leverage long-term dynamic information effectively. 
\begin{table}[!h]
    \centering
    \begin{tabular}{l | c}
        \toprule
        Model & Correlation with total number of snapshots  \\
        \midrule
        DySAT & 0.85 \\
        EGCN & -0.19 \\
        GCLSTM & -0.43 \\
        STGCN & 0.62 \\
        EdgeBank & 0.09 \\
        \bottomrule
    \end{tabular}
    \caption{Correlation between the number of snapshot and the difference in model performances ($\tau_\infty - \tau_1 $).}
    \label{tab:correlation_duration_performance}
\end{table}

\subsection{Discussion and future research direction} \label{sec:discussion}
These results highlight the necessity of analyzing the real capacity of dynamic graph learning models to capture temporal information. It is also essential to carefully select datasets to determine if a dynamic model is truly needed for future predictions. We demonstrated that several standard DTDGNNs models in the dynamic community are not consistently better at capturing temporal information compared to predicting solely from the previous snapshot (\cref{tab:all_temp_vs_tau1}). We also observed a lack of robustness concerning the size of the temporal receptive field (\cref{fig:tw}) and the inability of some models to handle dynamic information over long data sequences (\cref{tab:correlation_duration_performance}). Additionally, we found that a judicious choice of an optimal temporal context can significantly improve the performance of certain models (\cref{fig:tau_opt}).  We encourage the dynamic graph learning community to assess the true dynamic capability of models by integrating the proposed evaluations. 
\section{Conclusion}

In this work, we formalize the concept of the temporal receptive field and conduct an in-depth evaluation of it in the task of link prediction on DTDGs. We revealed the capacity of each model to capture dynamic information and pinpointed the datasets that require a long temporal context. Our analyses showed that having a very large temporal receptive field is surprisingly not synonymous with better predictions, and sometimes, not considering past information leads to better results.
\begin{credits}
\subsubsection{\discintname} All authors disclosed no relevant relationships.

\end{credits}
%
%
%
%
\clearpage
\bibliography{references}
\bibliographystyle{splncs04}

%
\clearpage
\setcounter{section}{0}
\renewcommand\thesection{\Alph{section}}
\section{Additional results}\label{sec:sup_results}
This \cref{tab:all_temp_vs_opt_tau} presents additional results, showing the comparison of Average Precision (AP) scores for various Dynamic Temporal Graph Neural Networks (DTGNNs) across multiple datasets. The results compare using all available temporal information $\tau_\infty$ versus an optimal temporal receptive field $\tau^*$. The average gains in AP scores for each model are also indicated.
\begin{table}[!h]
    \centering
    \begin{tabular}{c | c | lllll}
    \toprule
        $\tau$ & Datasets & EGCN \cite{pareja2019evolvegcn:graphs} & DySat \cite{sankar2018dysat:networks} & GCLSTM \cite{model_chen2022gclstm} & STGCN \cite{model_yu2017stgcn} & EdgeBank \cite{poursafaei2022towardsprediction}\\
        \midrule
         \multirow{12}{*}{$\tau_{\infty} $} & CanParl & 89.70 & 79.44 & 70.47 & 90.64 & 54.48\\
         & USLegis & 90.02 & 88.36 & 84.42 & 90.95 & 57.75\\
         &Trade & 91.95& 92.78 & 90.15 & 92.62 & 65.94\\
         & UCI-Message &54.33 & 77.04 & 87.22  & 82.14 & 76.28 \\
         &UNVote & 84.76 & 81.60 & 80.75 & 86.49 & 61.17\\
         &AS733 & 97.17 & 94.88 & 90.86 & 86.66 & 97.21\\
         &Enron & 91.58 & 91.84 & 91.26 & 83.73 & 79.35\\
         &Colab & 90.69 & 90.20 & 89.99 & 77.68 & 77.02\\
         &Bitcoin-OTC & 89.04 &  92.23 & 92.65 & 81.29 & 52.18\\
         &Bitcoin-Alpha & 76.19 & 81.10 & 65.91 & 70.72 & 57.43\\
         \midrule
         \multirow{12}{*}{$\tau^* $} & CanParl & 90.56 (1) & 83.63 (10) & 72.06 (6) & 90.65 (1) & 54.93 (3)\\
         & USLegis & 90.74 (5) & 90.38 (9) & 86.26 (2) & 90.95 (3) & 73.46 (1)\\
         &Trade & 92.12 (6)& 93.13 (2) & 91.02 (9) & 92.72 (6) & 82.42 (1)\\
         & UCI-Message & 84.85 (1) & 80.03 (9) & 87.22 ($\infty$) & 82.15 (2) & 76.28 ($\infty$)\\
         &UNVote & 86.96 (4) & 88.71 (4) & 84.57 (4) & 90.42 (2) & 97.30 (1)\\
         &AS733 & 97.37 (1)& 95.07 (2) & 98.03 (1) & 87.06 (6) & 97.21 ($\infty$) \\
         &Enron & 92.45 (1)& 91.84 ($\infty$) & 92.35 (5) & 84.49 (1) & 79.35 (7) \\
         &Colab & 91.01 (4) & 91.39 (1) & 90.19 (2) & 78.25 (7) & 77.02 (7)\\
         &Bitcoin-OTC & 89.99 (1) & 92.23 ($\infty$) & 92.65 ($\infty$) & 81.29 ($\infty$) & 52.18 ($\infty$)\\
         &Bitcoin-Alpha & 79.25 (2)& 81.10 ($\infty$) & 81.70 (1) & 70.75 (2) & 58.67 (8)\\
         \midrule
         & \cellcolor{slatecolor}\textbf{Avg gain} &\cellcolor{slatecolor} +3.40 &\cellcolor{slatecolor} +1.80 &\cellcolor{slatecolor} +3.24 &\cellcolor{slatecolor} +0.58 &\cellcolor{slatecolor} +1.92\\
         \bottomrule
         \bottomrule
    \end{tabular}
    \caption{Comparison of Average recision (AP) scores for various DTGNNs across multiple datasets, using all available temporal information $\tau_\infty$ versus an optimal temporal receptive field $\tau^*$. For $\tau^*$, we indicate in parentheses, the optimal value. }
    \label{tab:all_temp_vs_opt_tau}
\end{table}

\section{Datasets } \label{app:datasets}
Here, we describe the datasets used in our experiments. These descriptions are sourced from the Stanford SNAP database \cite{snapnets} as well as from the dynamic link prediction evaluation paper \cite{poursafaei2022towardsprediction}.
\begin{itemize}
    \item \textbf{CanParl}: Can. Parl. is a network that tracks how Canadian Members of Parliament (MPs) interacted between 2006 and 2019. Each dot represents an MP, and a line connects them if they both said "yes" to a bill. The line's thickness shows how often one MP supported another with "yes" votes in a year.
    \item \textbf{UsLegis}: USLegis is a Senate co-sponsorship network that records how lawmakers in the US Senate interact socially. The strength of each connection indicates how many times two senators have jointly supported a bill during a specific congressional session
    \item \textbf{Bitcoin-OTC}:  This is who-trusts-whom network of people who trade using Bitcoin on a platform called Bitcoin OTC. Since Bitcoin users are anonymous, there is a need to maintain a record of users' reputation to prevent transactions with fraudulent and risky users. Members of Bitcoin OTC rate other members in a scale of -10 (total distrust) to +10 (total trust) in steps of 1. This is the first explicit weighted signed directed network available for research \cite{snapnets}.
    \item \textbf{Bitcoin-Alpha}:  This is who-trusts-whom network of people who trade using Bitcoin on a platform called Bitcoin Alpha. Since Bitcoin users are anonymous, there is a need to maintain a record of users' reputation to prevent transactions with fraudulent and risky users. Members of Bitcoin Alpha rate other members in a scale of -10 (total distrust) to +10 (total trust) in steps of 1. This is the first explicit weighted signed directed network available for research \cite{snapnets}.
    \item \textbf{Trade}: UNTrade covers the trade in food and agriculture products between 181 nations over a span of more than 30 years. The weight assigned to each link within this dataset reflects the cumulative sum of normalized import or export values for agricultural goods exchanged between two specific countries.
    \item \textbf{UNVote}:  UNVote documents roll-call votes conducted in the United Nations General Assembly. Whenever two nations cast a "yes" vote for an item, the link's weight connecting them is incremented by one.
    \item \textbf{Contact}: Contact dataset provides insights into the evolving physical proximity among approximately 700 university students over the course of a month. Each student is uniquely identified, and links between them indicate their close proximity. The weight assigned to each link reveals the degree of physical proximity between the students
    \item \textbf{Enron}: Enron consists of emails exchanged among 184 Enron employees. Nodes represent employees, and edges indicate email interactions between them. The dataset includes 10 snapshots and does not provide node or edge-specific information
    \item \textbf{Colab}: Colab represents an academic cooperation network, capturing the collaborative efforts of 315 researchers from 2000 to 2009. In this network, each node corresponds to an author, and an edge signifies a co-authorship relationship.
    \item \textbf{UCI-Message}: This dataset is comprised of private messages sent on an online social network at the University of California, Irvine. Users could search the network for others and then initiate conversation based on profile information. An edge (u, v, t) means that user u sent a private message to user v at time t. The dataset here is derived from the one hosted by Tore Opsahl, but we have parsed it so that it can be loaded directly into SNAP as a temporal network \cite{snapnets}.

\end{itemize}
\end{document}